\documentclass[10pt, a4paper]{article}

\usepackage[final]{lrec2026} % this is the new style

\usepackage{pgfplots}
\usepackage{pgf}
\usepackage{tikz}
\usepackage{import} % for \import if needed
\usepackage{lmodern} % to avoid font errors
\usepackage{xspace}
\usepackage{amsmath}
\pgfplotsset{compat=1.18}
\usepackage{cleveref}
\usepackage{booktabs}
\usepackage{subcaption}
\newcommand{\npmi}{\texttt{NPMI}\xspace}
\newcommand{\twi}{\texttt{TWI}\xspace}
\newcommand{\twm}{\texttt{TWM}\xspace}
\newcommand{\umass}{\texttt{UMass}\xspace}
\newcommand{\lda}{\texttt{LDA}\xspace}
\newcommand{\nmf}{\texttt{NMF}\xspace}
\newcommand{\ttov}{\texttt{Top2Vec}\xspace}
\newcommand{\bertopic}{\texttt{BERTopic}\xspace}
\newcommand{\cfmf}{\texttt{CFMF}\xspace}
\newcommand{\cfmfe}{\texttt{CFMF-emb}\xspace}
\newcommand{\msize}[2]{\texttt{#1{\_#2}}\xspace}

\title{When Numbers Tell Half the Story: Human-Metric Alignment in Topic Model Evaluation
}

\name{%
  Thibault Prouteau$^{1}$, Francis Lareau$^{3, 4}$, Nicolas Dugué$^5$,\\ \large \textbf{Jean-Charles Lamirel$^{1,2}$,  Christophe Malaterre$^4$}%
}

\address{$^1$Université de Lorraine, CNRS, LORIA, Nancy (France) \\
$^2$Université de Strasbourg, Strasbourg (France) \\
$^3$Université de Sherbrooke, Dept of Philosophy and Applied Ethics, Sherbrooke (Canada)\\
$^4$Université du Québec à Montréal, Dept of Philosophy \& CIRST, Montréal (Canada)\\
$^5$Le Mans Université, LIUM, Le Mans (France) \\
\{thibault.prouteau, jean-charles.lamirel\}@loria.fr, 
\{francis.lareau, christophe.malaterre\}@uqam.ca,\\ nicolas.dugue@univ-lemans.fr
}

\abstract{
Topic models uncover latent thematic structures in text corpora, yet evaluating their quality remains challenging, particularly in specialized domains. Existing methods often rely on automated metrics like \emph{topic coherence} and \emph{diversity}, which may not fully align with human judgment. Human evaluation tasks, such as \emph{word intrusion}, provide valuable insights but are costly and primarily validated on general-domain corpora. This paper introduces \emph{Topic Word Mixing} (\twm), a novel human evaluation task assessing inter-topic distinctness by testing whether annotators can distinguish between word sets from single or mixed topics. \twm complements \emph{word intrusion}'s focus on intra-topic coherence and provides a human-grounded counterpart to diversity metrics. We evaluate six topic models--both statistical and embedding-based (\lda, \nmf, \ttov, \bertopic, \cfmf, \cfmfe)--comparing automated metrics with human evaluation methods based on nearly 4,000 annotations from a domain-specific corpus of philosophy of science publications. Our findings reveal that word intrusion and coherence metrics do not always align, particularly in specialized domains, and that \twm captures human-perceived distinctness while appearing to align with diversity metrics. We release the annotated dataset and task generation code. This work highlights the need for evaluation frameworks bridging automated and human assessments, particularly for domain-specific corpora.
 \\ \newline  }

\begin{document}

\maketitleabstract

\section{Introduction}

Topic models are unsupervised methods that aim to uncover latent thematic structures in document collections, representing topics as a distribution over words. Such methods can provide valuable insight into large corpora \cite{noichl2021modeling,malaterre2023emergence}. As topic models evolve--from classical \emph{Latent Dirichlet Allocation} (\lda, \citet{blei2003latent}) to embedding-based and neural variants--evaluating and comparing their quality remains a fundamental challenge.

The field has largely converged on two evaluation paradigms: automated metrics and human judgments. Automated metrics, such as \emph{topic coherence} and \emph{topic diversity} are most often based on word co-occurrence patterns, and enable low-cost comparisons across methods and hyperparameters. However, their relationship with human evaluations of topic quality remains debated \cite{hoyle2021automated}. Human evaluation methods, most notably \citet{changReadingTeaLeaves2009}'s \emph{word intrusion}, offer a valuable starting point by directly assessing topic coherence, though their validation has largely focused on general-domain corpora. This raises a two-sided question: \emph{to what extent do automated metrics capture the full range of human judgments, beyond coherence, and in particular in domain-specific contexts?}

A critical limitation of existing human evaluation tasks is their focus on \emph{intra-topic} coherence--whether words within a topic are semantically, or distributionally related, through the lens of \emph{word intrusion}. This provides an incomplete picture: a topic model can yield internally coherent topics that overlap significantly, failing to effectively partition the corpus. While automated diversity metrics attempt to measure \emph{inter-topic} distinctness, their relationship to human judgment remains unclear. Moreover, specialized domains that involve jargon and abstract concepts pose unique challenges. In academic fields, like philosophy of science, evaluating whether topics capture meaningful conceptual distinctions requires careful consideration of domain-characteristics, often falling back on domain experts.

This paper makes three main contributions. First, we introduce \emph{Topic Word Mixing} (\twm), a human evaluation task that assesses \emph{inter-topic} distinctness by testing whether annotators can identify if a word-set originates from a single topic or is a mixture of two topics. This task complements \emph{word intrusion}'s (\twi) focus on \emph{intra-topic} coherence acting like a human pendant to topic diversity, providing direct human judgment on topic boundaries.

Second, we conduct a comprehensive evaluation comparing six topic models (\lda, \nmf, \ttov, \bertopic, \cfmf, \cfmfe) using both automated metrics (coherence, diversity) and human evaluation (\emph{word intrusion}, \emph{topic word mixing}) totaling almost 4,000  annotations with 10 annotators on a domain-specific corpus spanning decades of philosophy of science publications. We release the dataset of annotated \emph{word intrusion} and \emph{topic word mixing} tasks, along with the code to generate such tasks, to foster the development of new evaluation metrics.

Third, we analyze the relationship between automated and human evaluation methods. Results reveal that \emph{word intrusion} and \emph{topic coherence} do not always align well, raising doubts about the reliability of coherence metrics in specialized corpora. This suggests that automated coherence measures based solely on domain-specific references may be insufficient and that broader, generalist contexts—such as those captured by large-scale embeddings—could improve evaluation. Consequently, these findings point to the need for developing embedding-based automated measures of coherence. Importantly, we find that constructing intrusion tasks in specialized domains requires additional considerations--particularly regarding the semantic similarity between intruder words and topic words--that have not been traditionally addressed. The human annotation results also indicate that \emph{word mixing} appears to track topic diversity reasonably well, at least in the present corpus. 

\Cref{sec:related_work} reviews related work on topic model evaluation. \Cref{sec:evaluation_framework} describes our corpus, automated metrics, \emph{word intrusion} and our newly introduced \emph{topic word mixing} human evaluations. \Cref{sec:experimantal_setup} details our experimental setup. \Cref{sec:results} reports results from both automated and human-centered evaluations. \Cref{sec:discussion} discusses implications and limitations, \Cref{sec:conclusion} concludes.

\section{Related Work}
\label{sec:related_work}

Evaluating topic models remains a fundamental challenge. Early work relied primarily on perplexity and likelihood measures as assessments of topic quality, which were shown to correlate poorly with human judgment \cite{blei2003latent, wallach2009evaluation}. \textit{Topic coherence} metrics were thus introduced, measuring the semantic similarity between top words within a topic. Several coherence variants have been introduced with diverse reference similarity corpus and aggregation functions. \npmi (\emph{Normalized Pointwise Mutual Information}) measures word co-occurrence patterns and was shown to correlate with human judgment \cite{newman2010automatic,lauMachineReadingTea2014}, while $C_v$ \citep{roder2015exploring} incorporates word vectors and sliding windows. \umass coherence uses document co-occurrence from the corpus itself. These metrics are widely used for model selection. 

Beyond coherence, researchers have proposed additional metrics capturing other aspects of topic quality. Notably, \emph{topic diversity} measures the proportion of unique words across topics, as a way to identify topic overlap and redundancy \cite{dieng2020topic,tran2013topic}. \emph{Coverage} metrics measure how well topics represent the entire range of documents present in the corpus. However, the relationship between these metrics and human perception of topic quality is not established. Technical vocabulary and abstract concepts can present a challenge for automated metrics.

Given the limitations of automated metrics, human evaluation remains essential for validating topic model quality. \citet{changReadingTeaLeaves2009} introduced two influential tasks: \emph{word intrusion} (\twi) and \emph{topic intrusion}. In \twi, annotators identify an intruder among a topic's top words; success correlates with human perceived coherence. In \emph{topic intrusion}, annotators are asked to identify which topic does not fit a given document. When conducted on two generalist corpora, \emph{New York Times} and \emph{Wikipedia} documents, these tasks revealed that models optimized for likelihood demonstrate less interpretable topics for humans.

Subsequent work has extended these evaluation paradigms. \citet{newman2010automatic} investigated topic quality by asking human to rate topics. \citet{lauMachineReadingTea2014} have extended and automatized \emph{word intrusion} comparing it to automated coherence metrics, finding moderate but imperfect correlation; more recent work proposed hybrid methods bridging human and machine evaluation \cite{yangLLMReadingTea2025}. Such paradigms have also been used to assess the interpretability of word embeddings \cite{murphyLearningEffectiveInterpretable2012b, faruquiSparseOvercompleteWord2015,subramanianSPINESParseInterpretable2018a,prouteau2022embedding}. However, these studies have predominantly focused on topic coherence and on general-domain corpora (news, Wikipedia), leaving open questions about evaluation and correlation of human-judgment with other automated metrics in domain-specific settings.

\section{Evaluation Framework}\label{sec:evaluation_framework}
\subsection{Corpus}
The corpus represents an extensive and comprehensive collection of the full texts from eight leading philosophy of science journals, spanning from 1931--the publication date of the earliest journal--through 2017 (\textit{British Journal for the Philosophy of Science}, \textit{European Journal for Philosophy of Science}, \textit{Erkenntnis}, \textit{International Studies in the Philosophy of Science}, \textit{Journal for General Philosophy of Science}, \textit{Philosophy of Science}, \textit{Studies in History and Philosophy of Science Part A} and \textit{Synthese}).

With a total of 16,917 articles and approximately 65 million word occurrences (23.7k types), this corpus includes both English and non-English items (6\% of primarily German, Dutch, and French). Following \citet{malaterre2022early}, foreign language documents were machine-translated into English. No further preprocessing was required by embedding-based methods. The input for remaining methods was tokenized, part-of-speech tagged and tokens were lemmatized.  Low-frequency terms were removed by excluding words that appeared in fewer than 50 sentences across the corpus. To focus topic induction on semantically salient lexical items, only content words (nouns, verbs, adjectives, adverbs) were retained to produce vector representations needed by other topic models, such as word‑frequency Term $\times$ Document Matrix (TDM); normalized TF‑IDF TDM; and Doc2Vec vectors.      

\subsection{Topic Models}

Our goal is to evaluate topics across different modeling paradigms. We selected methods involving diverse approaches: probabilistic (\emph{LDA}, \emph{NMF}, \emph{CFMF}), and embedding-based (\emph{BERTopic}, \emph{CFMF-emb}, \emph{Top2Vec}).
\textbf{Latent Dirichlet Allocation} (LDA) \citep{blei2003latent} treats each document as a mixture of topics, where each topic is a distribution over words, estimated via Gibbs sampling on a word-frequency term-document matrix (TDM).
\textbf{CFMf} combines Feature Maximization \citep{lamirel2016new}, which selects topic-terms using an F-measure scheme on normalized TF-IDF vectors, with an adapted version of Growing Neural Gas clustering \citep{fritzke1994growing} using prototype renormalization \cite{lamirelold}  and exploiting normalized TF-IDF document vectors as input. \textbf{CFMf-emb} uses the same Feature Maximization but applies GNG with renormalization on normalized transformer-derived document embeddings.
\textbf{Top2Vec} \citep{angelov2020top2vec} embeds words and documents using Doc2Vec \citep{le2014distributed}, discovers clusters via HDBSCAN \citep{campello2013density} without prespecified \emph{K}, and extracts top terms via cosine proximity to cluster centroids.
\textbf{BERTopic} \citep{grootendorst2022bertopic} uses HDBSCAN on transformer embeddings, and applies class-based TF-IDF to rank words per topic.
\textbf{Nonnegative Matrix Factorization} (NMF) \citep{lee1999learning} factorizes a  TF-IDF matrix into nonnegative term-topic and topic-document matrices via alternating least squares.

Models were built with \emph{K} = 10, 25, and 50 topics. For \lda, \cfmf, \cfmfe, and \nmf, \emph{K} was set explicitly. For \ttov and \bertopic, effective K was indirectly controlled by tuning HDBSCAN's minimum cluster size; this resulted in models with \emph{K} = 10, 25, 48 for \ttov and \emph{K} = 10, 25, 54 for \bertopic. Document embeddings are extracted (except for Top2vec that uses Doc2Vec) using \texttt{stella\_en\_1.5B\_v5} \cite{zhang2025jasperstelladistillationsota}, due to its performances for long texts (131,072+ tokens) \citep{muennighoff-etal-2023-mteb}. All models produce ranked \emph{top-W} word descriptors to describe each topic.

\subsection{Automated Metrics}

\textbf{Coherence} measures the interpretability of a topic by quantifying the semantic relatedness of its top words; higher coherence indicates that top terms tend to co-occur in the reference corpus. Coherence metrics include \emph{PMI}, \emph{NPMI}, \emph{U-Mass}, and \emph{Coherence $C_V$}. Here we used \emph{Coherence $C_V$} because it is supposed to correlate well with human judgments on topic quality \citep{roder2015exploring}. Using a sliding window of size $s$ (typically 110), $C_V$ builds for each top word a smooth NPMI-based context vector over the set of $N$ top words. It then aggregates these vectors into a topic vector, computes cosine similarity between each word vector and the topic vector (via segmentation), and reports the mean cosine across all words and topics of the corpus under investigation. The $C_V$ score ranges from $[-1, 1]$, typically normalized to $[0, 1]$, with 1 indicating more coherent topics.

Topic \textbf{diversity} quantifies the distinctness of topics by measuring the overlap of their top terms; higher diversity means topics share fewer common top words. Given $K$ topics and $N$ top words for each topic, let $TopW_k$ be the set of top $N$ words of topic $k$ and $U = \bigcup_{k=1}^K TopW_k$ the set of unique top words across all topics. Topic diversity is: $\mathbf{TD=\frac{|U|}{K\cdot N}}$, it ranges from 0 (all topics share the same top words) to 1 (no top word is shared between topics).
\label{sec:diversity}

\subsection{Human Evaluation}

Human evaluation is critical for assessing topic model quality, though scarcely implemented compared to automated metrics. \citet{changReadingTeaLeaves2009}'s seminal work introduced \emph{word intrusion} and \emph{topic intrusion} tasks. \citet{mimno2011optimizing} proposed coherence measures correlating with human judgments, while \citet{roder2015exploring} systematically compared automated coherence metrics against human evaluations. \citet{lauMachineReadingTea2014,yangLLMReadingTea2025} have revisited these paradigms and provide classifier and LLM-based approaches of \emph{word intrusion} and \emph{topic coherence rating}. While coherence metrics have been shown to correlate with human judgment across multiple studies, \citet{hoyle2021automated} demonstrated that this alignment is imperfect and inconsistent. Furthermore, most validation work has been conducted on general-domain corpora, raising questions about generalization to specialized domains.

\paragraph{(Topic) Word Intrusion.}

(\twi) was originally introduced in \citet{changReadingTeaLeaves2009}. Aimed at evaluating the coherence of topics, \twi adopts a contrastive approach by presenting the top ranked words (head) of a topic along with an intruder taken randomly, following certain conditions. The motivating idea is that a topic is coherent if humans are able to discriminate the intruder hidden among the top words of the topic. In cases where topics are hard to interpret, evaluators will typically choose intruders at random, which implies a low coherence for the topics presented. Although \cite{lauMachineReadingTea2014} found a correlation between human judgment in word intrusion and \npmi on classical corpora (\emph{New York Times}, \emph{Wikipedia}), \citet{hoyle2021automated} have suggested that this correlation may not hold consistently across datasets or experimental setups.

Word intrusion has been frequently used to assess the quality of topic models. However, to the best of our knowledge, word intrusion tests have mainly been conducted on generalist corpora and rarely, if at all, on domain-specific corpora. One of the main obstacles to word intrusion evaluation is the cost of human annotations. This partly explains why researchers favor automated metrics. In addition, when expert knowledge is required, the search for annotators can be even more difficult.

Despite these obstacles, we conducted a \emph{word intrusion} evaluation on the present domain-specific corpus to investigate how human evaluation of topic coherence relates to automated metrics in this context.

\paragraph{Topic Word Mixing.} As noted above, topic diversity provides a complementary metric, reflecting how distinct topics are from one another. Unlike coherence, there is currently no human evaluation framework for assessing topic diversity. Automated measures of diversity typically rely on lexical overlap or similarity between topic-word distributions, making diversity an intrinsic property of the model rather than a perceptual one. We therefore propose to ground quantitative diversity metrics in human judgment, by viewing topic distinctness as a human-perceived proxy for topic diversity.

We thus introduce \textit{\textbf{Topic Word Mixing}} (\twm), a human evaluation task aimed at assessing inter-topic distinctness. The task probes whether human annotators can determine if a given set of words originates from a single topic or from a mixture of two. Intuitively, if a topic model learns thematically diverse topics, humans should easily recognize mixed sets as heterogeneous. Conversely, when topics overlap semantically, this distinction becomes more difficult to make.

Formally, let $\mathcal{T} = \{\beta_1, \dots, \beta_K\}$ be a topic model and 
$W_i^{(m)}$ denote the set of the top-$m$ words of topic $i$ as ordered in $\beta_i$. 
For an even integer $n$, a \emph{topic word mixing task} of size $n$ is defined as
$$
\mathcal{M}(n) =
\begin{cases}
\big(W_i^{(n/2)} \cup W_j^{(n/2)}, \; y=2\big), & i \neq j, \\[6pt]
\big(W_k^{(n)}, \; y=1\big), & 
\end{cases}
$$

\noindent where $y \in \{1,2\}$ is the class label: $y=1$ indicates a single-topic set of $n$ words, 
and $y=2$ indicates a mixture of two topics with $n/2$ words each.

This formulation allows for a direct human-based assessment of topic distinctness, thus providing an interpretable measure of how separable the topics are in semantic space.

\section{Experimental Setup}\label{sec:experimantal_setup}

\paragraph{Automated Metrics} We chose to measure topic model quality with two metrics: \emph{coherence} $C_V$ \cite{roder2015exploring} and \emph{diversity} (\cref{sec:diversity}). While these metrics enable efficient comparison across models and hyperparameters, their alignment with human judgment--particularly in specialized domains--remains an open question that our human evaluation tasks aim to address.

\paragraph{(Topic) Word Intrusion.} In the \twi evaluation we conducted here, each word intrusion task consisted of \textbf{5 words}: the 4 top-words of a given topic (the head) and 1 intruder word. The intruder was selected according to the following criteria:
\begin{itemize}
    \item It belongs to the lower 50\% of the evaluated topic's ranked words while appearing in the top 10\% of words in at least one other topic. This ensures that the intruder is semantically distinct but still plausible in the corpus.
    \item Its average frequency in the corpus had to match that of the evaluated topic's top words. This constraint ensures that annotators could not rely on word frequency as a cue.
\end{itemize}
To verify annotator attentiveness, we included simple control tasks in which the intruder word was highly salient. Finally, the word order within each task was randomized.

\label{sec:word_intrusion_setup}
\paragraph{Topic Word Mixing.} To assess topic distinctness, we propose to focus the \twm tasks on the most confusable topic pairs within each model. Such pairs can be identified by computing the cosine similarity between topic embeddings, where each topic embedding corresponds to the mean vector of the top 50 topic words encoded here with the \texttt{stella\_en\_1.5B\_v5} model. For each topic, the most similar counterpart is selected to form candidate pairs for the \twm task. This procedure directs the evaluation toward subtle, fine-grained distinctions between semantically proximate topics rather than trivially distinct ones.

Each \twm consisted of \textbf{8 words}. For mixed-topic tasks involving 2 topics, these 8 words consisted of the top 4 words of each topic; for single-topic tasks, the 8 words consisted of the top 8 words of the sampled topic. The number of words (8) was chosen to balance informativeness and cognitive feasibility by providing enough context for judgment, while remaining close to short-term memory limits \cite{miller1956magical}. Half of the tasks presented a mixture of two topics, and half presented words from a single topic, ensuring a balanced classification setting.

To mitigate cognitive load and ensure that the task primarily tested topic-level distinctness rather than annotator ability for grouping terms, half of the words in each task were bolded: in single-topic tasks, these bolded words were randomly chosen; in mixed-topic tasks, one of the two topics was represented by bold words. This visual cue relieves participants from inferring topic clusters and focuses their effort on determining whether the set contains one or two topics. 

Word order within each task was randomized to prevent participants from relying on adjacency cues. For each model, topic pairs were sampled uniformly among the most similar pairs, ensuring diversity of topics across the evaluation set. Easily solvable control tasks were included to check for annotator attentiveness.

\label{sec:word_mixing_setup}

\paragraph{Annotators.} Ten annotators—undergraduate philosophy students with training in epistemology and philosophy of science—were recruited to evaluate the topic models using \twi and \twm. All possessed relevant  knowledge in the corpus domain and received financial compensation for their participation. The annotation process was distributed across the annotators, with five individuals assigned to \twi and five to \twm. To mitigate fatigue and maintain annotation quality, each task was divided into four tracks, allowing annotators to take breaks between segments.

\begin{figure}
    \centering
    \includegraphics[width=\linewidth]{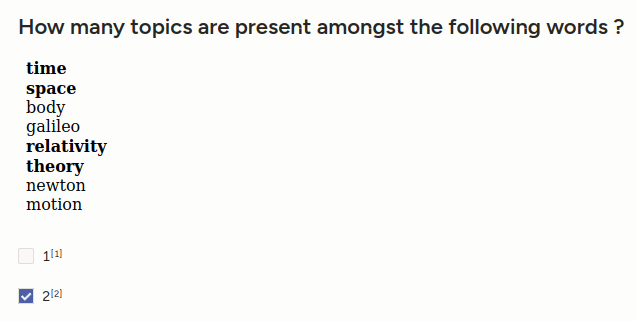}
    \caption{Example of a two-topic Word Mixing task in \emph{LabelStudio}.}
    \label{fig:TWM_labelstudio}
\end{figure}

\paragraph{Annotation Platform \& Track Descriptions.} The annotation tasks were conducted using LabelStudio's online platform \cite{LabelStudio}. Annotators accessed four separate tracks for each task, with all items randomized prior to presentation to prevent order effects. For the word intrusion task (\twi), annotators were shown five words and asked to identify the single intruder word that did not belong with the others.
The word mixing task (\twm) interface presented \cref{fig:TWM_labelstudio} has the annotators select "1" or "2" perceived topics.

\paragraph{Human Evaluation Dataset.} Based on these criteria, a total of 18 models—six topic modeling methods trained with approximately 10, 25 and 50 topics each) were evaluated using both \twi and \twm. For \twi, all topics from models of size 10 were annotated, while for models of size 25 and 50, 60\% of topics were randomly sampled for annotation. This resulted in 330 distinct tasks, amounting to 1650 annotations across five annotators. For \twm, both single-topic and mixed-topic tasks needed to be sampled. Accordingly, all models with 10 topic were fully included, while 50\% of tasks were uniformly randomly sampled for models with 25 and 50 topics. In total, 467 tasks were evaluated across the 18 models, yielding 2335 annotations.
The code and annotation interface used to generate and sample intrusion and mixing tasks from topic models are open-sourced on Github\footnote{\url{https://github.com/thibaultprouteau/LREC_2026_Prouteau_Human_Metrics_Topic_Models}}.

\section{Results}\label{sec:results}

\paragraph{Automated Metrics.}

\Cref{tab:coherence_diversity} details the coherence and diversity score for each model to be annotated in \twi and \twm. 
Regarding coherence, \ttov achieves the best results regardless of the granularity of the topic model, ranging from $0.72$ to $0.80$ depending on $K$. Feature maximization-based methods (\cfmf, \cfmfe) follow with slightly more moderate coherence ($0.67$-$0.72$) while \nmf shows lower scores ($0.59$-$0.67$). \bertopic and \lda exhibit the weakest coherence, with \lda scoring the lowest across all configurations ($0.47$-$0.55$). 

Concerning topic diversity, \ttov leads again, maintaining near-perfect scores ($0.98$-$1$), indicating minimum overlap in top words across topics. \cfmf, \cfmfe, and \nmf show relatively high diversity for models with 10 topics ($0.90$-$0.96$) but experience notable decline with a greater number of topics, dropping to $0.73$-$0.81$ for $K=50$. This pattern--decreasing diversity with increasing $K$--is consistent across models, suggesting that as the number of topics grows, they tend to exhibit greater overlap. \lda and \bertopic show moderate diversity ($0.66$-$0.80$). Despite being based on embeddings like \cfmfe, \bertopic does not produce highly diverse topics.

\begin{table}[h]
\begin{tabular}{@{}l||ccc||ccc@{}}
\hline
          & \multicolumn{3}{c}{Coherence-10}             & \multicolumn{3}{c}{Diversity-10}              \\ \hline
$K\approx$      & 10            & 25            & 50            & 10            & 25            & 50            \\ \hline
\ttov     & \textbf{0,72} & \textbf{0,79} & \textbf{0,80} & \textbf{1,00} & \textbf{1,00} & \textbf{0,98} \\
\cfmf     & 0,69          & 0,72          & 0,72          & 0,96          & 0,85          & 0,81          \\
\cfmfe    & 0,67          & 0,69          & 0,70          & 0,90          & 0,79          & 0,73          \\
\nmf      & 0,59          & 0,65          & 0,67          & 0,90          & 0,88          & 0,78          \\
\bertopic & 0,52          & 0,58          & 0,62          & 0,79          & 0,70          & 0,66          \\
\lda      & 0,47          & 0,54          & 0,55          & 0,80          & 0,74          & 0,66       \\ \hline   
\end{tabular}
\caption{Coherence $C_V$ and diversity (10 words) of each topic model.}
\label{tab:coherence_diversity}
\end{table}

\paragraph{Topic Word Intrusion.} The \emph{inter-annotator agreement} reaches an average of 80.3\% and a Fleiss' Kappa \cite{fleiss1973equivalence} of $0.688$ which indicates a \emph{substantial agreement} \cite{landis1977measurement}.

\begin{figure}[t]
    \centering
    \includegraphics[width=\linewidth]{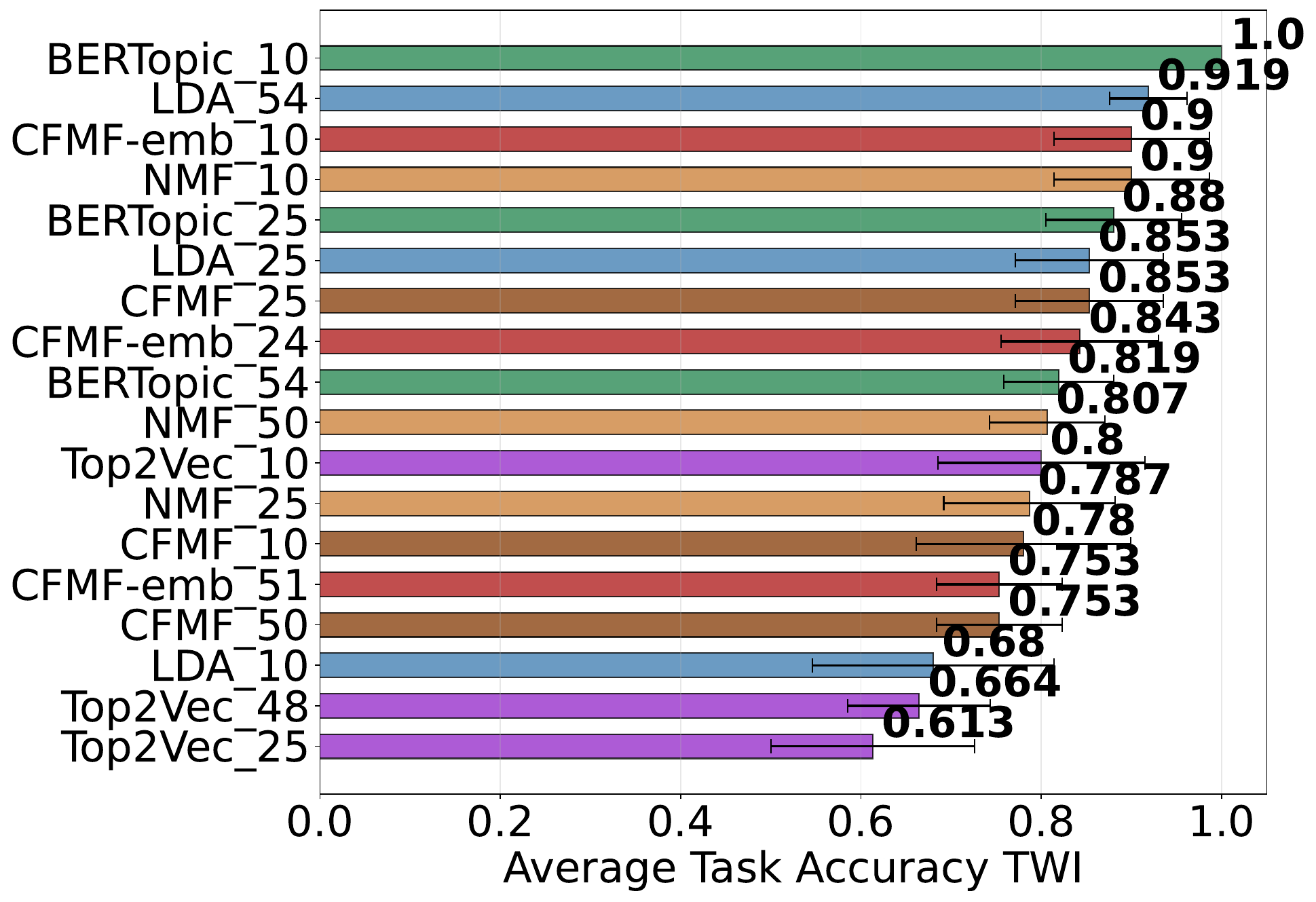}
    \caption{Macro average annotation accuracy for \twi  per model (95\% confidence intervals).}
    \label{fig:twi_models}
\end{figure}

\Cref{fig:twi_models} presents the average of annotation task success rates across all model configurations (model\_K). Model performance varies considerably within model families depending on model size $K$. \msize{\bertopic}{10} achieves perfect accuracy (1.00) while \msize{\bertopic}{54} drops to 0.82.  Similarly, \msize{\cfmfe}{10} reaches 0.90, but \msize{\cfmfe}{51} falls to 0.75. This pattern of higher accuracy for smaller models seems to hold for most models (except \msize{lda}{10}), indicating that annotators find topics more coherent when $K$ is small. However, \msize{\lda}{54} outperforms many models at similar topic numbers, and \ttov consistently underperforms regardless of $K$, with \msize{\ttov}{25} scoring lowest at 0.61. It is worth noting that confidence intervals remain large despite the overall substantial agreement between annotators. Yet, \ttov is subpar, in comparison to \msize{\bertopic}{10} and \msize{\lda}{54}.

\begin{figure}[t]
    \centering
    \includegraphics[width=\linewidth]{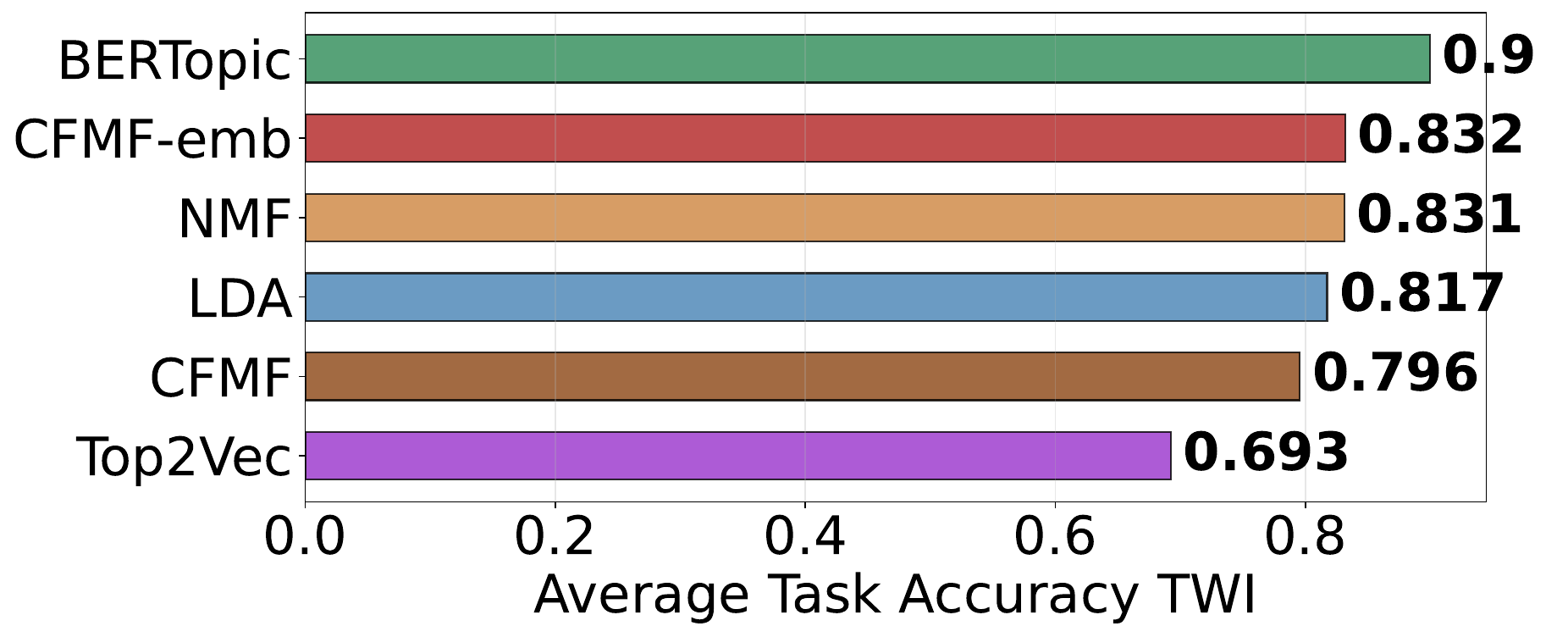}
    \caption{Macro average annotation accuracy for \twi per model family.}
    \label{fig:twi_family}
\end{figure}

The aggregated results per model family are presented in \Cref{fig:twi_family}. As shown, \bertopic achieves the highest performance on \twi(0.900) followed by \cfmfe, \nmf and \lda. The \cfmf variant, which does not use embeddings, attains a lower accuracy (0.796) compared to its embedding-based counterpart. Finally, \ttov yields the lowest overall performance among all model families (0.693).

\begin{figure}[b]
    \centering
    \begin{subfigure}{0.5\textwidth}
        \centering
        \includegraphics[width=\linewidth]{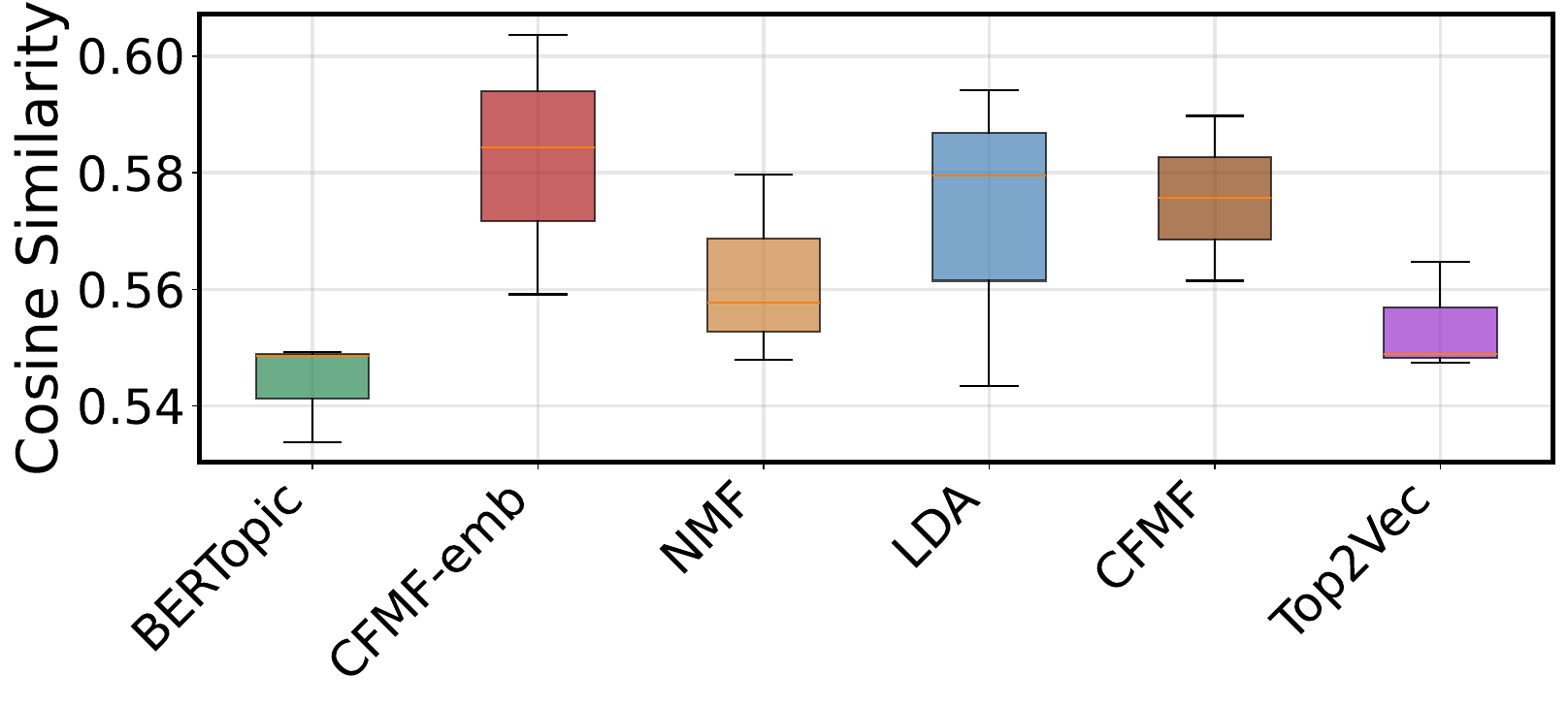}
        \caption{Cosine similarity between topic top-4 words and intruder for \twi per model family.}
        \label{fig:distri_cosine}
    \end{subfigure}%
    \hfill%
    \begin{subfigure}{0.5\textwidth}
        \centering
        \includegraphics[width=\linewidth]{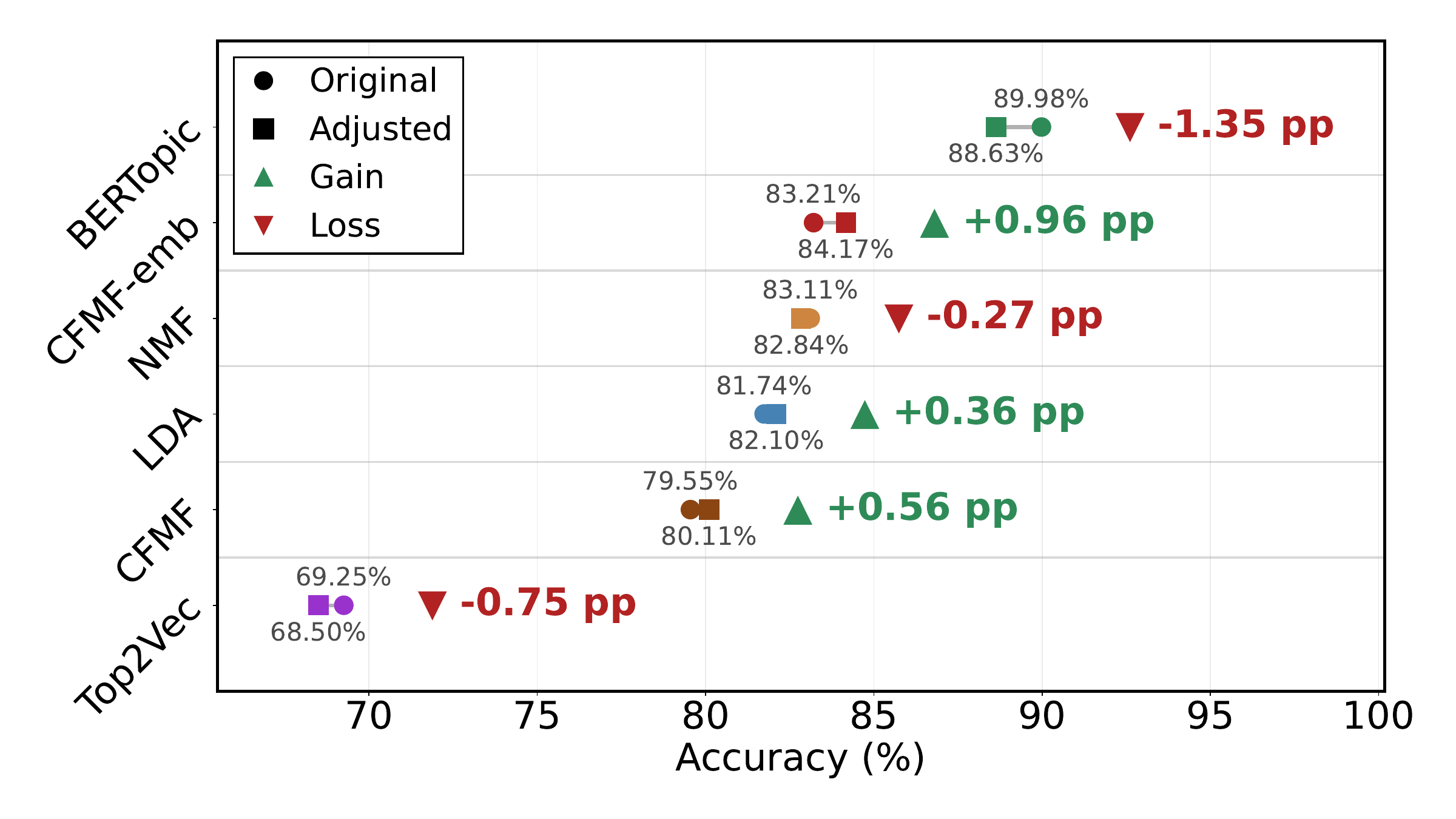}
        \caption{Adjustment of macro average annotation accuracy scores for \twi per model family taking into account the similarity between intruders and topic top-4 words.}
        \label{fig:adj_difficulty}
    \end{subfigure}
    \caption{Adjusting results to account for task difficulty (intruder-top-words similarity).}
\end{figure}

Traditionally, \twi tasks are constructed by choosing  words from the topic top-terms and, most importantly, by selecting an intruder based on word ranking constraints. However, relying solely on word ranking for intruder selection appears imperfect. Indeed, a posteriori examination of the distribution of average cosine similarity between top words and intruder (computed using embeddings from \texttt{stella\_en\_1.5B\_v5}), reveals clear discrepancies, as depicted in \Cref{fig:distri_cosine}. This matters since, intuitively, if an intruder is semantically close to the topic top terms, the task should become more difficult. In this respect, models such as \bertopic and \ttov are likely advantaged, whereas \cfmfe may have received more challenging tasks overall. Average task accuracy is inversely correlated with the mean similarity between top-words and intruder ($r=-0.145, p=0.0079$).

Results are then adjusted, based on the residuals of tasks to the fitted slope, to correct for task difficulty. In the adjusted results \cref{fig:adj_difficulty}, \cfmfe, \lda and \cfmf gain accuracy by alleviating the impact of intruder to topic head similarity. \nmf and \lda decrease slightly and \bertopic remains the leader, but loses 1.35 percentage point of accuracy. Though such adjustment likely corrects for task difficulty, the overall ranking of model families does not change here.  

When comparing \twi annotation accuracy with coherence $C_V$, we can see that $\ttov$ human perceived coherence doesn't align with automated measurement. A similar finding holds for \bertopic that has comparatively lower $C_V$ coherence but achieves higher accuracy \twi. Thus, automated co-occurrence based coherence measures may not fully capture human interpretability, at least not as well in specialized domains as in more generic corpora. This divergence underscores the complimentary nature of human evaluation to automated metrics in topic modeling.

\paragraph{Topic Word Mixing}

\begin{figure}[h]
    \centering
    \includegraphics[width=\linewidth]{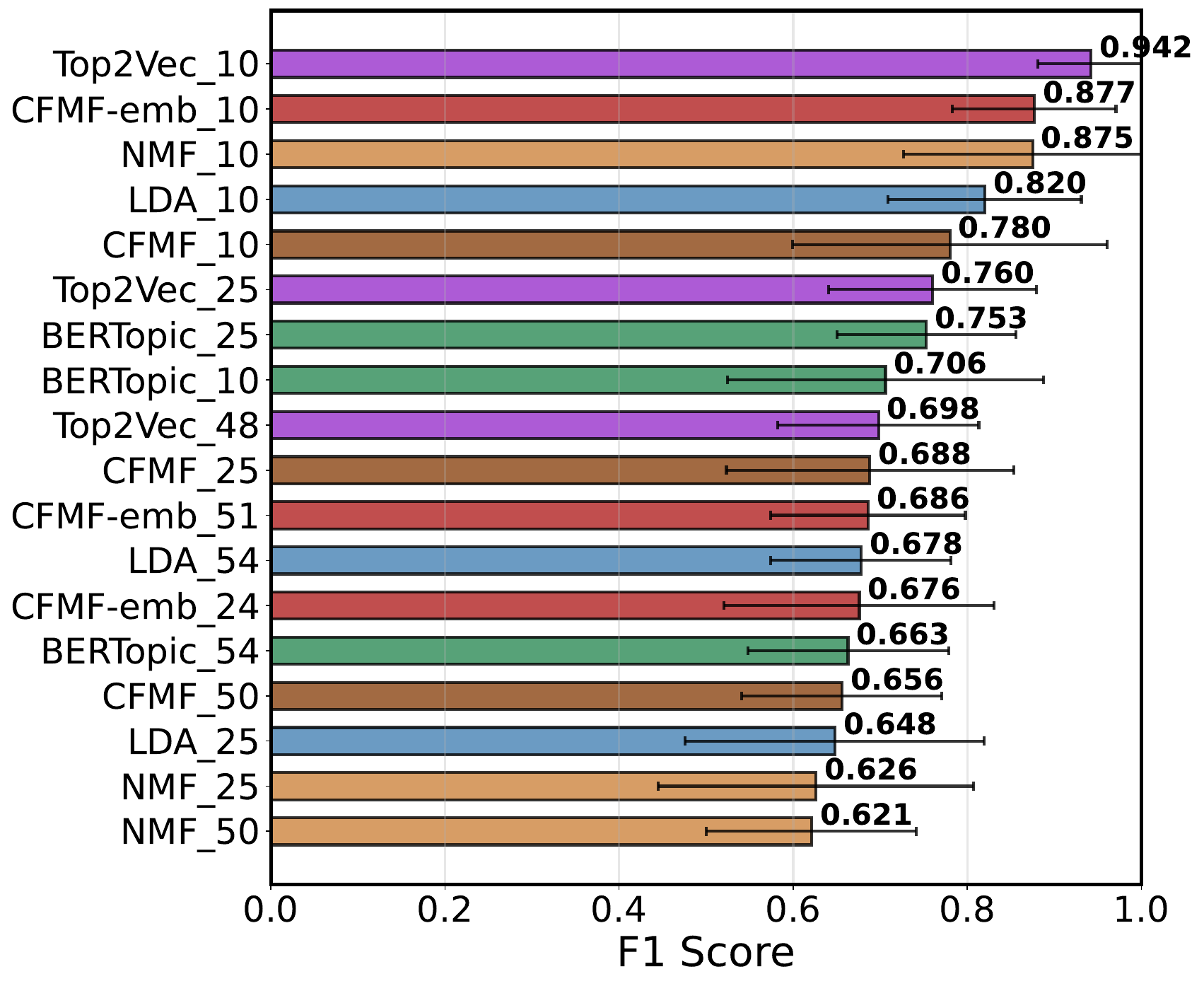}
    \caption{Macro average task F1 Score for \twm per model (with 95\% confidence intervals).}
    \label{fig:f1_models}
\end{figure}

The \emph{inter-annotator agreement} for \twm reaches an average of 77.1\% and a Fleiss' Kappa of 0.270, which is a \emph{fair agreement} and is a testament to the difficulty of the task. Indeed, we chose to mix topics that are the closest for which the decision boundary may be thin. \Cref{fig:f1_models} presents the mixing results per model. \ttov maintains good performances across granularities (0.698-0.942) with \msize{\ttov}{10} scoring highest. The \msize{\cfmfe}{10} and \msize{\nmf}{10} variants follow closely (0.87). \bertopic and \cfmf exhibit more variability across model sizes. Confidence intervals remain significant for some models for which detecting two-topic tasks was lower. 

\begin{figure}[h]
    \centering
    \includegraphics[width=\linewidth]{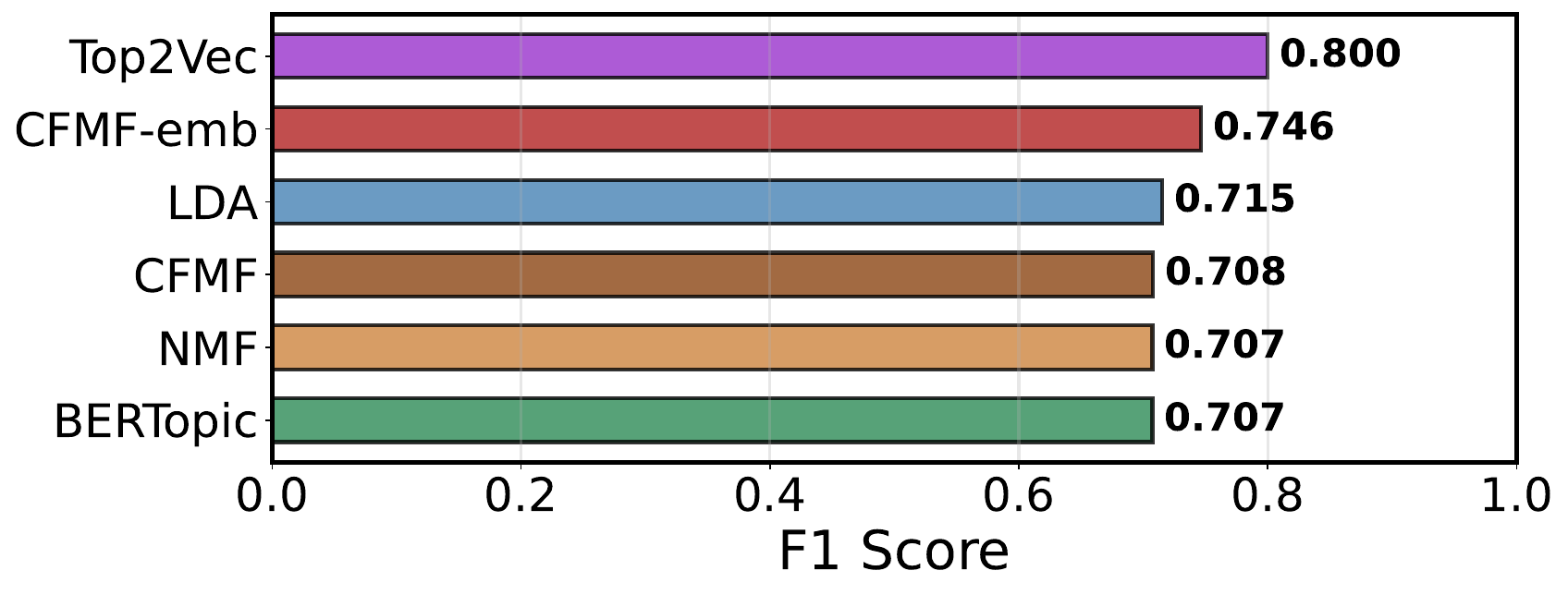}
    \caption{Macro average task F1 Score for \twm per model family.}
    \label{fig:f1_families}
\end{figure}

When considering families of models (\Cref{fig:f1_families}), \ttov achieves the best score (0.800) followed by \cfmfe (0.746). \lda, \cfmf, \nmf and \bertopic obtain similar results.

Compared with automated measures of \emph{diversity}, the \twm results show a consistent pattern. \ttov has near perfect diversity scores (0.98-1.00) and outperforms all other models on \twm. Interestingly, while \cfmfe shows lower quantitative diversity than \cfmf, it achieves higher human-perceived diversity. Nonetheless, the contrast that was apparent between models in terms of diversity (\cref{tab:coherence_diversity}) seems less pronounced in the \twm evaluation. 

Note that the \twm results also align with coherence measures for \ttov and \cfmfe. This suggests that \emph{Topic Word Mixing} may complement \emph{Topic Word Intrusion} in more ways than just adding an inter-topic dimension. Indeed, for two mixed topics to be perceived as distinct, each must first be internally coherent; conversely, when a task involves a single topic whose words lack coherence, it becomes more difficult for annotators to recognize it as such. Thus, \twm indirectly captures aspects of both topic distinctness and internal coherence, reinforcing its value as a complementary human evaluation method.

Another more general remark is that when considering \twi and \twm scores jointly and producing an average ranking of the methods based on that combination, \cfmfe distinguishes itself from others. 

\section{Discussion}\label{sec:discussion}

\paragraph{Misalignment between automated metrics and human judgment.} Our results reveal misalignment between automated coherence and human annotation with the \emph{word intrusion} on this domain-specific corpus. \ttov achieved the highest coherence ($C_V$=0.72-0.80) and near perfect diversity (0.98-1.00), yet performs poorest on \twi (0.693). Conversely, \bertopic excels at \twi (0.900) despite having the second-lowest automated coherence (0.52-0.62). These discrepancies suggest that co-occurrence metrics may not reliably capture human-perceived coherence in specialized domains, where technical vocabulary and abstract concepts pose unique challenges.

\paragraph{Performance under joint human evaluation.} Large confidence intervals limit firm conclusions about model performance. With this caveat in mind, \cfmfe achieves the best overall performance (mean rank = 2) with balanced performance across both tasks. \nmf and \lda follow (mean rank = 4 and 3.5) with moderate performance. \cfmf ranks lower (mean rank = 4.5), suggesting potential benefits from incorporating word embeddings. Finally, \bertopic and \ttov show somewhat different profiles, with \bertopic appearing to favor coherence and \ttov appearing to favor diversity, though both show discrepancies with their respective metric predictions.

\paragraph{What does topic word mixing capture?} The \twm task appears to capture topic distinctness, at least partially aligning with the diversity metric. However, it likely probes richer aspects of topics. We hypothesize that \twm indirectly captures both diversity and some coherence: if topics are incoherent, distinguishing mixed top-words becomes difficult; conversely, coherent and distinct topics yield clearer mixing judgments. 

\paragraph{Granularity effects remain unclear.} Although we relied on annotators with domain expertise, our annotations provide insufficient evidence to draw strong conclusions about the effects of model granularity ($K$) on human-perceived quality (e.g. \msize{\lda}{54} outperforms \msize{\lda}{10}). Far more annotations would be required to establish clear trends. Consequently, the choice of $K$ should be guided by research goals and human appraisal. For instance, studies seeking to map the general thematic structure of a corpus—such as the overall landscape of philosophy of science—may be adequately served by models with relatively few topics (e.g. 10–25). Conversely, investigations targeting narrowly defined themes, such as specific philosophical theories of scientific explanation, may require models with a larger number of topics to ensure that at least one captures the phenomenon of interest.

\paragraph{Task difficulty considerations.} Task difficulty should be considered when designing human-based evaluations. Our a posteriori analysis revealed that intruder-to-topic similarity (measured via word embedding) significantly affects \twi. Future work should control for such similarity a priori to ensure balanced task difficulty across models. This would help prevent confounding task difficulty with  model quality.

\paragraph{Toward embedding-based measures.} The embedding-based similarity we use to assess task difficulty creates an opportunity to revisit prior evaluations \cite{aletras2013evaluating} and to develop new coherence metrics grounded in distributional semantics. Traditional coherence relies on corpus-internal-coherence (whether it is measured on the corpus at hand or a more general corpus). While they are appropriate for capturing domain-specific patterns, they are limited by corpus boundaries. Large pre-trained embeddings encode broader world knowledge from vast corpora, potentially offering richer semantic representations. An externalized coherence measure based on embedding similarity between topic words could complement corpus-internal metrics like $C_V$ coherence leveraging the broader embedded lexical and conceptual knowledge. Since our annotators possess, to some degree world knowledge and domain expertise, embedding may operationalize coherence in ways that better align with human judgment.

\paragraph{Scope limitations.} Coherence and diversity likely capture only two aspects of topic quality. Models can yield topics that are both highly coherent and clearly distinct, yet fail to represent the variety of topics present in the corpus by focusing on narrow, salient themes. For this reason, coverage measures--assessing how well topics capture the full range of documents--should complement coherence and diversity evaluations, whether they are metric or human-based. Unfortunately, coverage is difficult to assess qualitatively through human evaluation due to the usual scale of corpora. Consequently, quantitative coverage measures remain indispensable complements to human-centered evaluations.

\section{Conclusion}\label{sec:conclusion}

Our results challenge the assumption that automated topic coherence metrics and human word intrusion evaluations are strongly correlated. In our domain-specific setting, coherence scores (e.g., $C_V$) and word intrusion accuracy often diverged; standard metrics may not reliably reflect human perception of coherence in specialized fields. This calls for caution when relying solely on automated coherence measures, especially outside general-domain corpora. To complement word intrusion, we introduced \emph{Topic Word Mixing} (\twm), a new human evaluation probing inter-topic distinctness. \twm offers a direct, human-centered view of topic diversity, complementing diversity metrics. Our results show that automated diversity metrics partially align with human judgments. Looking forward, our study underscores the importance of combining multiple evaluation perspectives--automated and human-centered--to obtain a more complete picture of topic model quality. We release our annotated datasets of almost 4,000 examples and our code to support further research in this direction. Future work should explore embedding-based coherence measures to better align with human evaluations.

\section{Acknowledgements}
J.-C.L. \& T.P. acknowledge funding from AID-RAPID SUMINO. F.L. acknowledges funding from Canada Social Sciences and Humanities Research Council (Postdoctoral Fellowships 756-2024-0557). C.M. acknowledges funding from Canada Social Sciences and Humanities Research Council (Grant 430-2018-00899) and Canada Research Chairs (CRC-950-230795). N.D. acknowledges support from  Agence Nationale de la Recherche (ANR) through the DIGING project (ANR-21-CE23-0010). The authors acknowledge that they were granted a free academic license of “Label Studio” by HumanSignal, available at \url{https://labelstud.io}. Experiments presented in this paper were carried out using the Grid'5000 testbed, supported by a scientific interest group hosted by Inria and including CNRS, RENATER and several Universities as well as other organizations (see \url{https://www.grid5000.fr}).

\section{Bibliographical References}\label{sec:reference}

\bibliographystyle{lrec2026-natbib}
\bibliography{bibliography}

\end{document}